\documentclass{article}
\usepackage[T1]{fontenc}

\usepackage{microtype}
\usepackage{graphicx}
\usepackage{booktabs} 
\usepackage{xurl}
\usepackage{float}
\usepackage{hyperref}
\usepackage{tikz}
\usepackage{amsmath}
\usepackage{amssymb}
\usepackage{bm}
\usepackage{enumitem}
\usepackage{wrapfig}
\usepackage{needspace}

\usepackage[final]{corl_2026}
\hypersetup{
  pdftitle={InCoM: Intent-Driven Perception and Structured Coordination for Mobile Manipulation},
  pdfauthor={Jiahao Liu, Wenbo Cui, Zhongpu Xia, Yongliang Wang, Haoran Li, Dongbin Zhao}
}
\makeatletter
\renewcommand{\@conferencelocation}{Austin, Texas, USA}
\makeatother

\title{InCoM: Intent-Driven Perception and Structured Coordination for Mobile Manipulation}

\author{
    \textbf{Jiahao Liu$^{1,2,\dagger}$, Wenbo Cui$^{1,\dagger}$, Zhongpu Xia$^{3}$,} \\
    \textbf{Yongliang Wang$^{1}$, Haoran Li$^{1,2,*}$, Dongbin Zhao$^{1,2}$} \\
    \\
    \small $^1$SKL-MAIS, Institute of Automation, Chinese Academy of Sciences \\
    \small $^2$School of Advanced Interdisciplinary Sciences, University of Chinese Academy of Sciences \\
    \small $^3$Anyverse Dynamics \\
    \small $^\dagger$Equal contribution. \quad $^*$Corresponding author \\
    \small \texttt{\{liujiahao2077, yongliangwang1997\}@gmail.com}, \texttt{xiazhongpu5@163.com} \\
    \small \texttt{\{cuiwenbo2023, lihaoran2015, dongbin.zhao\}@ia.ac.cn}
}

\begin{document}
\maketitle

\begin{abstract}
Mobile manipulation is a fundamental capability for general-purpose robotic agents, requiring both coordinated control of the mobile base and manipulator and robust perception under dynamically changing viewpoints. However, existing approaches face two key challenges: strong coupling between base and arm actions complicates control optimization, and perceptual attention is often poorly allocated as viewpoints shift during mobile manipulation.
We propose \textbf{InCoM}, an intent-driven perception and structured coordination framework for mobile manipulation. InCoM infers latent motion intent to dynamically reweight multi-scale perceptual features, enabling stage-adaptive allocation of perceptual attention. To support robust cross-modal perception, InCoM further incorporates a geometric--semantic affinity refinement mechanism that enhances multimodal correspondence. On the control side, we design a decoupled coordinated flow matching action decoder that explicitly models bidirectional base--arm interactions, alleviating optimization difficulties caused by control coupling.
Under matched non-privileged observations and task-specific training data, InCoM improves mean success over DSPv2 by \textbf{33.0}, \textbf{26.1}, and \textbf{23.6 percentage points} on SetTable, TidyHouse, and PrepareGroceries, respectively. Furthermore, its effectiveness is consistently validated in real-world mobile manipulation tasks, where InCoM maintains a superior success rate over existing baselines. The project website is available at \href{https://liujiahao2077.github.io/InCoM.github.io/}{\nolinkurl{liujiahao2077.github.io/InCoM.github.io}}.
\end{abstract}


\section{Introduction}
\label{intro}
Mobile manipulation is a fundamental capability for realizing general-purpose robotic policies in real-world environments and has attracted increasing attention from both academia and industry~\cite{mobile_aloha, ac-dit, brs, dspv2}. Compared with tabletop manipulators or mobile bases, mobile manipulation requires simultaneous control of the robotic arm and the mobile base, as well as continuous decision-making and control in complex, dynamic, and partially observable environments~\cite{harmonic, simultaneousperceptioninteraction}. A key challenge is generating stable coordinated actions between the mobile base and the manipulator. Due to inevitable execution errors in real robots, small deviations in base motion can be amplified at the manipulator end-effector under decoupled or weakly coordinated control~\cite{brs}, leading to degraded manipulation accuracy or task failure. Introducing conditional dependencies between the mobile base and the manipulator at the action decoding stage~\cite{ac-dit, brs} can partially alleviate this issue. However, such joint action modeling typically represents coordination as a one-way dependency, limiting the model's ability to capture the bidirectional coupling and mutual compensation inherent in real-world tasks.

\begin{figure}[!t]
\begin{center}
\centerline{\includegraphics[width=\textwidth]{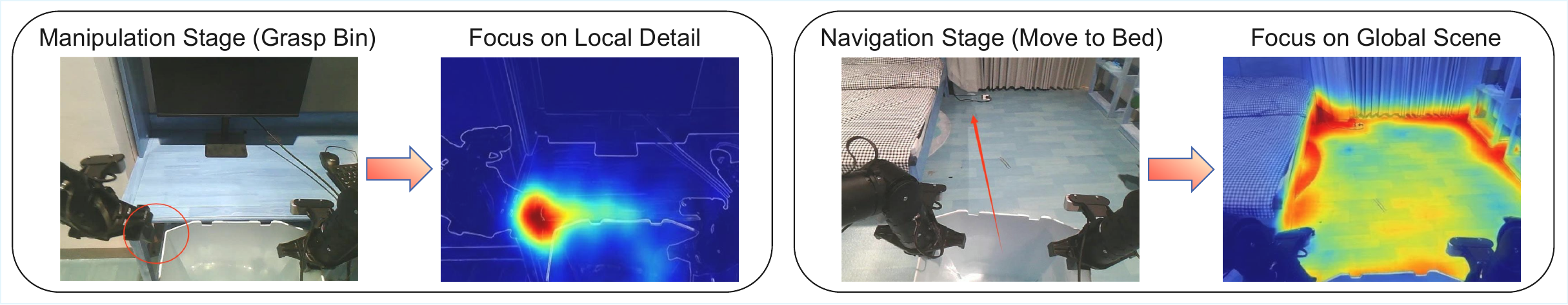}}
\caption{
\textbf{Dynamic perceptual attention during mobile manipulation.}
The left half is the color image, and the right half is a schematic diagram of perceptual attention.
During manipulation, perceptual attention is primarily focused on local interaction targets; for example, the agent should attend to whether the robotic arm has successfully grasped the trash bin (left). During navigation, perceptual attention shifts toward understanding the global structure to identify free-space (right).
}
\label{fig:sketch_map}
\end{center}
\end{figure}

Another important yet often overlooked challenge is the allocation of perceptual information under dynamically changing viewpoints, which we refer to as the dynamic perceptual attention problem. Unlike single-mode settings such as tabletop manipulation or navigation, mobile manipulation combines and switches among multiple operational modes with distinct perceptual requirements.
For example, in a task such as \emph{moving a trash bin next to the bed} shown in Fig.~\ref{fig:sketch_map}, the policy must focus on the interaction location between the gripper and the rim of the trash bin for safe and reasonable grasping, whereas during navigation toward the bed, attention should shift to free-space regions and obstacle boundaries to avoid collisions. However, most existing end-to-end policies~\cite{dspv2, act, dp, 3d_dp} neglect this issue, making it difficult for optimization to capture the evolving perceptual demands as the task progresses. As a result, perceptual attention becomes dispersed during execution, ultimately degrading task success rates.

To address these challenges, we propose an end-to-end framework named InCoM for mobile manipulation that jointly considers stage-adaptive perception and coordinated action generation. InCoM explicitly models the evolution of perceptual attention throughout task execution, while enabling stable and efficient generation of coordinated base--arm actions. Specifically, to address the dynamic perceptual attention problem induced by viewpoint changes, we propose a multi-scale perception architecture that infers the robot's latent motion intent and dynamically allocates perceptual attention according to the current task stage, enabling adaptive fusion of global scene context and local detail cues. This design improves robustness to motion-induced viewpoint changes. In addition, to better integrate information across modalities, we introduce a cross-modal alignment and fusion module that models geometric consistency and semantic correspondence, thereby enhancing multimodal information fusion. Finally, to tackle joint action generation, we design an action generation mechanism that enables information interaction between the mobile base and the manipulator during decoding, allowing bidirectional coordination and mutual compensation. The main contributions are summarized as follows:

\begin{itemize}[leftmargin=*, itemsep=2pt, topsep=0pt, parsep=0pt] 
\item We propose {InCoM}, an end-to-end framework for mobile manipulation that jointly models intent-driven perception and coordinated action generation within a unified learning formulation, alleviating perceptual dispersion and control coupling.
\item We design an {Intent-Driven Pyramid Perception Module (IDPPM)} that infers latent motion intent to adaptively reweight multi-scale perceptual features, enabling stage-aware allocation between global context and local fine-grained details. We further propose a {Dual-stream Affinity Refinement Module (DARM)} that explicitly models and fuses geometric and semantic affinities to enhance cross-modal alignment.
\item We introduce a {Decoupled Coordinated Flow Matching (DCFM)} decoder that explicitly captures bidirectional coordination between the mobile base and the manipulator, producing smooth and expressive actions.
\item We demonstrate success-rate gains of \textbf{33.0/26.1/23.6 percentage points} over DSPv2 under matched training data and non-privileged observations across three ManiSkill-HAB~\cite{mshab} scenarios. Its effectiveness is further validated in real-world mobile manipulation tasks, where it consistently maintains a superior success rate over existing baselines.
\end{itemize}

\section{Related Work}
\subsection{Mobile Manipulation Policies}
Early mobile manipulation systems coordinated navigation and manipulation through phase-switching rules or high-level planners~\cite{integratedtaskmotionplanning, doasican, ok-robot, robomatrix, wildLma}. LOAT~\cite{loat} uses LLM-derived object affinities for object-goal navigation, while LeAffordNav~\cite{leaffordnav} combines LLM-guided exploration with affordance-aware navigation for open-vocabulary mobile manipulation. RoboGPT~\cite{robogpt} integrates LLM-based subgoal planning, navigation and manipulation skills, and feedback-driven replanning. Although effective in structured settings, modular designs can suffer from discontinuities and error accumulation under tightly coupled phases or dynamic environments. HoMeR~\cite{homer} advances hierarchical mobile manipulation by integrating policy learning with whole-body control.

End-to-end imitation learning uses chunk-wise Transformers~\cite{act}, multi-modal visual fusion~\cite{dspv2}, or diffusion-based trajectory modeling~\cite{m2_diffuser} to improve temporal coherence and reduce hand-designed rules. These methods usually represent base and arm actions as a unified vector without explicitly modeling their mutual dependencies. Pretrained robot policies and vision--language--action (VLA) models~\cite{rt-2, octo, openvla, openvla-oft, vla-survey} enhance generalization, with QDepth-VLA~\cite{qdepth-vla} adding auxiliary quantized-depth prediction for manipulation. However, many focus on tabletop tasks without fully capturing movement--manipulation dependencies. To model base--arm dependencies, BEHAVIOR Robot Suite~\cite{brs} uses hierarchical diffusion decoding, while AC-DiT~\cite{ac-dit} conditions whole-body actions on mobility-related context. Such unidirectional modeling struggles to capture reciprocal coordination. InCoM instead exchanges information between base and arm branches during decoding, explicitly modeling bidirectional coordination to improve action coherence and execution stability.

\subsection{Multi-Modal and Multi-Scale Visual Perception}
Mobile manipulation requires global perception for navigation and fine-grained local perception for interaction~\cite{ac-dit}. Pretrained 2D models provide semantic cues~\cite{clip, dinov2, siglip-2}, while 3D methods capture spatial structure and object geometry~\cite{minkowski, lift3d, pointtransformerv3}, including metric distance and free space not explicitly encoded by DINOv2 semantics. For geometry-based manipulation, GAPartManip~\cite{gapartmanip} provides material-diverse data and actionable part-pose annotations, while DiffuDepGrasp~\cite{diffudepgrasp} models depth-sensor noise for sim-to-real grasping. Beyond modality choice, fixed-scale representations~\cite{ac-dit} limit adaptation of perceptual granularity across task stages. Multi-scale features capture local details in shallow layers and global context in deeper layers~\cite{fromcliptodino}. Despite their success in detection and scene parsing~\cite{fpn, deeplab, pyramidsceneparsingnetwork, pointnet++, swintransformer, mask3d}, stage-dependent multi-scale scheduling remains underexplored in mobile manipulation. InCoM uses action history and current DINOv2 features to adaptively reweight its feature hierarchy.

Cross-modal fusion requires correspondence between 3D point clouds and 2D image features. Geometric projection~\cite{pointpainting} is sensitive to calibration errors and projection noise, while Transformer-based methods model alignment through cross-attention~\cite{transfusion} or 3D positional embeddings~\cite{petr, dspv2}. CLAR~\cite{cl3r} combines masked 3D reconstruction, contrastive alignment, and local deformable attention through dedicated pretraining. DARM instead models geometric and semantic affinities in parallel, decoupling spatial correspondence from semantic matching. It adaptively balances geometric consistency and semantic relevance without additional large-scale cross-modal pretraining, improving cross-modal alignment robustness.

\section{Methodology}
\subsection{Problem Formulation and Overall Framework}
Mobile manipulation can be modeled as a discrete-time partially observable Markov decision process (POMDP). At time step $t$, the robot receives an observation 
$\mathcal{O}_t = \{P_t, \mathcal{I}_t, \mathbf{s}_t\}$, where $P_t$ denotes a sparse 3D point cloud, $\mathcal{I}_t = \{I_t^v\}_{v=1}^V$ represents multi-view RGB images, and $\mathbf{s}_t$ is the proprioceptive state. The action is factorized into mobility and manipulation subspaces: $\mathbf{a}_t = [\mathbf{a}_t^{base}, \mathbf{a}_t^{arm}] \in \mathcal{A}$, where $\mathbf{a}_t^{base} \in \mathbb{R}^{d_{base}}$ controls the mobile base and $\mathbf{a}_t^{arm} \in \mathbb{R}^{d_{arm}}$ controls the manipulator and associated joints. Our objective is to learn a conditional policy $\pi_\theta(\mathbf{a}_{t:t+T_p-1} | \mathcal{O}_t, \mathbf{a}_{t-H:t-1})$ that generates coordinated whole-body actions over the next $T_p$ steps from the current observation and action history.

\begin{figure}[!t]
\centering
\includegraphics[scale=0.45]{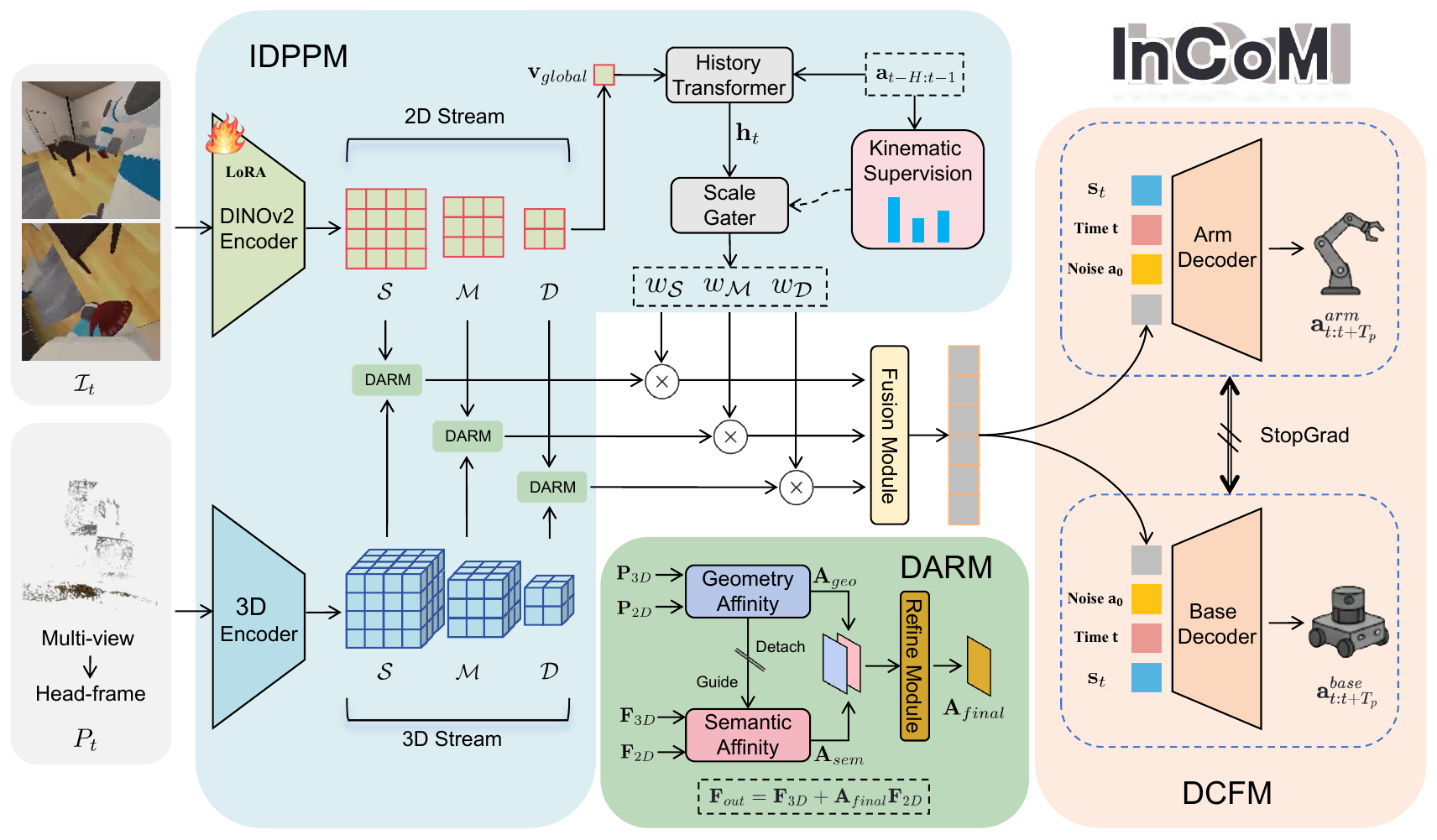}
\caption{\textbf{Overview of InCoM.} 
The framework integrates intent-driven multi-scale perception (IDPPM), dual-stream cross-modal affinity refinement (DARM), and decoupled flow-based action generation (DCFM) to produce coordinated mobile manipulation.
}
\label{fig:framework}
\end{figure}

We introduce InCoM, a mobile manipulation framework consisting of an Intent-Driven Pyramid Perception Module (IDPPM), a Dual-stream Affinity Refinement Module (DARM), and a Decoupled Coordinated Flow Matching (DCFM) decoder, as shown in Fig.~\ref{fig:framework}. Multi-modal observations are encoded into a multi-scale feature pyramid, where DARM models geometric and semantic affinities between point cloud and image features for cross-modal alignment. IDPPM infers implicit motion intent from historical actions and global context to reweight features across scales for stage-aware perception allocation. The resulting features are decoded by DCFM to generate coordinated actions for the mobile base and manipulator.

\subsection{Intent-Driven Pyramid Perception Module}
In mobile manipulation, navigation and manipulation require perceptual information at different spatial scales. However, existing policies~\cite{ac-dit, brs, dspv2} typically adopt fixed-scale perception encoders, limiting their ability to adapt between global scene understanding and local fine-grained modeling. To address this, we propose IDPPM (Intent-Driven Pyramid Perception Module), which dynamically reweights multi-scale features using historical actions and global visual context.

\textbf{Hierarchical Feature Pyramid.} We extract features using a sparse 3D encoder~\cite{minkowski} and a pretrained 2D visual encoder (DINOv2~\cite{dinov2}), and align them across three abstraction levels $k \in \{ \mathcal{S}, \mathcal{M}, \mathcal{D} \}$. The shallow, mid, and deep levels respectively capture local details for fine-grained manipulation, intermediate spatial context, and global scene structure for mobile base navigation.

\textbf{Intent Modulation.} To dynamically allocate perceptual attention according to task stages, we introduce a History Transformer, which maps the historical action sequence $\mathbf{a}_{t-H:t-1}$ and the current global DINOv2 feature $\mathbf{v}_{global}$ into an implicit intent vector $\mathbf{h}_t$. Here, intent is a history-derived phase estimate rather than an explicit future goal. Subsequently, an MLP-based ScaleGater network $\phi$ maps $\mathbf{h}_t$ to normalized hierarchical weights $\mathbf{w}_t$:
\begin{equation}
\mathbf{w}_t = [w_{\mathcal{S}}, w_{\mathcal{M}}, w_{\mathcal{D}}] = \text{Softmax}(\phi(\mathbf{h}_t))
\end{equation}

\textbf{Auxiliary Kinematic Supervision.} To align perceptual focus with the robot's kinematic state, we introduce a kinematics-guided loss to regularize feature allocation. Specifically, we quantify the activity levels of the mobile base and the manipulator by computing the $L2$ norms of their action increments over the historical action sequence:

\begin{equation}
I_{\mathrm{base}} \!=\! \frac{1}{T_{h}} \sum_{t} \| \Delta \mathbf{a}_{t}^{\mathrm{base}} \|_{2}, \,
I_{\mathrm{arm}} \!=\! \frac{\alpha}{T_{h}} \sum_{t} \| \Delta \mathbf{a}_{t}^{\mathrm{arm}} \|_{2}
\end{equation}
where $T_h=H$ is the history length and $\alpha=0.375$ balances the different dimensionalities and magnitudes of base/arm activity. The target weight distribution $\mathbf{w}^*$ is defined as:
\begin{equation}
w^*_{\mathcal{D}} \propto I_{base}, \quad w^*_{\mathcal{S}} \propto I_{arm}, \quad w^*_{\mathcal{M}} \propto \sqrt{ w^*_{\mathcal{S}} w^*_{\mathcal{D}}}
\end{equation}
By minimizing the KL divergence $\mathcal{L}_{scale} = D_{KL}(\mathbf{w}_t || \mathbf{w}^*_t)$, the model learns to emphasize deep, global features during rapid motion and shallow, local features during fine-grained manipulation. To prevent overfitting to rigid kinematic variations, we further introduce an entropy regularization term:
\begin{equation}
\mathcal{L}_{scale} = D_{KL}(\mathbf{w}_t || \mathbf{w}^*_t) + \lambda_{ent} \cdot \sum_{k \in \{\mathcal{S, M, D}\}} w_k \log w_k
\end{equation}
where $\lambda_{ent}$ is the entropy regularization coefficient, and $k \in \{\mathcal{S}, \mathcal{M}, \mathcal{D}\}$ indexes the shallow, middle, and deep pyramid levels. $w_k$ denotes the predicted weight of the $k$-th level.

\subsection{Dual-stream Affinity Refinement Module}
Cross-modal alignment requires both geometric consistency and semantic correspondence between color images and point clouds. Existing methods~\cite{dspv2} often entangle positional and semantic information within a single attention computation. We propose a Dual-stream Affinity Refinement Module (DARM), which separately models geometric affinity and semantic affinity and fuses them through a lightweight mechanism.

\textbf{Dual-Stream Affinity Modeling.}
Let $\mathbf{F}_{3D}, \mathbf{P}_{3D} \in \mathbb{R}^{N \times D}$ denote 3D voxel features and positional encodings, and $\mathbf{F}_{2D}, \mathbf{P}_{2D} \in \mathbb{R}^{M \times D}$ denote 2D patch features and positional embeddings.

Geometric affinity $\mathbf{A}_{geo}$ is a scaled dot product between encodings of 3D voxel centers and fixed 2D patch-grid positions augmented with camera identity; semantic affinity $\mathbf{A}_{sem}$ is the corresponding dot product between voxel and image features. Computing them in distinct representation spaces enables decoupled modeling of geometric and semantic relationships. DARM does not explicitly use camera intrinsics or extrinsics internally: ``geometric'' refers to its positional inputs, not a calibrated reprojection. Batched dot products construct dense $N\times M$ maps without materializing an $N\times M\times D$ pairwise feature tensor. The maps are concatenated and adaptively fused through a grouped $1\times1$ refiner $f_{\text{refine}}$, followed by patch-wise softmax and image-value aggregation. Token counts are bounded at each scale; implementation and memory costs are detailed in Appendix~\ref{policy_details}.

\textbf{Geometry-Guided Regularization.}
Since geometric relationships are typically more stable during training, we further use geometric affinity to regularize semantic matching. Specifically, $\mathbf{A}_{geo}$ is normalized to construct a transport cost matrix $\mathbf{C} = \mathbf{1} - \text{Softmax}(\mathbf{A}_{geo})$, based on which the Sinkhorn--Knopp algorithm produces a soft correspondence distribution $\mathbf{T}^*$. This distribution represents geometrically plausible cross-modal associations.

We encourage the semantic attention distribution $\mathbf{P}_{sem} = \text{Softmax}(\mathbf{A}_{sem})$ to align with this geometric prior by minimizing the KL divergence:

\begin{equation}
\mathcal{L}_{align} = D_{KL}(\mathbf{P}_{sem} \parallel \mathrm{sg}(\mathbf{T}^*)),
\end{equation}

where $\mathrm{sg}(\cdot)$ denotes the stop-gradient operator. Geometric affinity acts as a stable prior, while detaching the correspondence target prevents semantic-matching gradients from updating the geometric branch through this loss. Sinkhorn uses five iterations with $\epsilon=0.05$.

\subsection{Decoupled Coordinated Flow Matching}
Mobile manipulation requires handling high-dimensional action spaces while preserving cross-part coordination. Existing methods~\cite{ac-dit, brs} typically rely on unidirectional dependency modeling, which fails to capture bidirectional interactions among body parts. We propose DCFM (Decoupled Coordinated Flow Matching), a decoupled yet coordinated action decoder based on conditional flow matching, reducing iterative sampling overhead while improving action smoothness and stability.

\textbf{Decoupled Generation Architecture.} We first factorize the high-dimensional whole-body action space into base actions $\mathbf{a}^{base}$ and arm actions $\mathbf{a}^{arm}$. To avoid the error accumulation and computational overhead of diffusion models, we construct a direct linear path between noise $\mathbf{z}_0 \sim \mathcal{N}(\mathbf{0},\mathbf{I})$ and an expert action sequence $\mathbf{z}_1$. We denote flow time by $\tau$ to distinguish it from environment time $t$:
\begin{equation}
\mathbf{z}_\tau = (1-\tau)\mathbf{z}_0 + \tau\mathbf{z}_1, \quad \tau \in [0,1]
\end{equation}
The corresponding constant velocity is $\mathbf{u}=\mathbf{z}_1-\mathbf{z}_0$, which is predicted by $v_\theta$. Separate Transformer decoders are trained for the mobile base and the manipulator to minimize:
\begin{equation}
\mathcal{L}_{flow} = \sum_{k \in \{base, arm\}} \| v_\theta^k(\mathbf{z}_\tau^k, \tau, c) - (\mathbf{z}_1^k - \mathbf{z}_0^k) \|^2
\end{equation}
where $c = \psi(\mathcal{O}_t, \mathbf{a}_{t-H:t-1})$ is the multimodal conditioning embedding. All reported results use one Euler integration step at inference.

\textbf{Explicit Bidirectional Coordination.} While decoupled generation alleviates the dimensionality challenge, it overlooks physical coupling between body parts. To address this, we introduce an information interaction mechanism between the two generation streams. As shown in Fig. \ref{fig:action_head}(c) in Appendix~\ref{policy_details}, each decoder branch maintains a learnable Trend Token that aggregates predicted motion trends. At each Transformer layer, the mobile base and the manipulator exchange this information via cross-attention with stop-gradient to ensure bidirectional coordination.

\subsection{Overall Optimization Objective}
The final training objective is a weighted combination of the flow matching loss, the auxiliary kinematic supervision, and the geometry-guided attention regularization:
\begin{equation}
\mathcal{L}_{total} = \mathcal{L}_{flow} + \lambda_{scale} \mathcal{L}_{scale} + \lambda_{align} \mathcal{L}_{align}
\end{equation}

\section{Experiments}
We systematically evaluate the effectiveness of the InCoM framework in mobile manipulation tasks. First, in Section \ref{exp:mshab_eval}, we conduct experiments on three different mobile manipulation scenarios in the ManiSkill-HAB~\cite{mshab} simulation platform: SetTable, TidyHouse, and PrepareGroceries. The results show that InCoM consistently outperforms existing state-of-the-art baselines in terms of task success rates. Next, in Section \ref{exp:ablation}, we perform a comprehensive ablation study to assess the contribution and necessity of each component of InCoM. In addition, in Section \ref{exp:real_world}, we further validate the effectiveness of InCoM through real-world robot experiments.

\subsection{ManiSkill-HAB Evaluation}
\label{exp:mshab_eval}

\textbf{Evaluation Settings.} We evaluate InCoM on ManiSkill-HAB~\cite{mshab} under perception-constrained settings. To better reflect real-world conditions, reproduced methods exclude simulator-privileged signals and rely on visual inputs and robot proprioception. InCoM and all reproduced baselines use the same 1,000 RL-generated demonstrations per task; representations remain method-specific. We compare against DP~\cite{dp}, ACT~\cite{act}, DSPv2~\cite{dspv2}, WB-VIMA~\cite{brs}, and the pretrained $\pi_0$~\cite{pi0}. AC-DiT~\cite{ac-dit} results are taken from its paper, since its code was unavailable when our comparison was completed; these are unmatched, privileged-observation results. Appendix~\ref{details_of_mshab} gives the full protocol and baseline configurations.

\begin{table}[ht]
\caption{SetTable success rates (\%, mean$\pm$SD). Bold marks the highest mean in each column. $\dagger$: results from the AC-DiT paper using privileged object/goal and EE-to-base poses and a grasp indicator. $\ddagger$: pretrained robot policy.}
\label{table:set_table}
\begin{center}
\begin{small}
\setlength{\tabcolsep}{2.2pt}
\begin{tabular}{lcccccccc}
\toprule
Method & \shortstack{Pick\\Apple} & \shortstack{Place\\Apple} & \shortstack{Open\\Fridge} & \shortstack{Pick\\Bowl} & \shortstack{Place\\Bowl} & \shortstack{Open\\Drawer} & \shortstack{Close\\Drawer} & Mean \\
\midrule
DP      & $0.5\!\pm\!0.5$ & $54.5\!\pm\!5.1$ & $63.0\!\pm\!4.0$ & $2.1\!\pm\!0.9$ & $63.5\!\pm\!4.3$ & $5.3\!\pm\!1.8$ & $89.4\!\pm\!1.9$ & 39.8 \\
ACT     & $1.6\!\pm\!0.9$ & $21.2\!\pm\!3.2$ & $74.6\!\pm\!3.3$ & $9.0\!\pm\!1.9$ & $21.7\!\pm\!2.9$ & $48.1\!\pm\!3.7$ & $91.5\!\pm\!1.8$ & 38.2 \\
WB-VIMA & $1.6\!\pm\!1.4$ & $57.7\!\pm\!4.7$ & $27.0\!\pm\!3.7$ & $1.6\!\pm\!1.8$ & $60.3\!\pm\!4.2$ & $5.3\!\pm\!1.4$ & $87.3\!\pm\!2.8$ & 34.4 \\
DSPv2   & $1.4\!\pm\!0.3$ & $65.2\!\pm\!3.6$ & $73.4\!\pm\!3.2$ & $1.4\!\pm\!0.8$ & $\bm{85.7\!\pm\!2.4}$ & $29.9\!\pm\!3.2$ & $98.4\!\pm\!1.4$ & 50.8 \\
$\pi_0^{\ddagger}$ & $26.6\!\pm\!3.8$ & $48.9\!\pm\!5.2$ & $90.5\!\pm\!1.9$ & $26.6\!\pm\!4.5$ & $56.8\!\pm\!5.9$ & $75.5\!\pm\!2.5$ & $88.7\!\pm\!2.0$ & 59.1 \\
AC-DiT$^{\dagger}$ & $33.3\!\pm\!1.9$ & $33.3\!\pm\!9.4$ & $\bm{90.7\!\pm\!5.0}$ & $36.0\!\pm\!6.5$ & $17.3\!\pm\!6.8$ & $81.3\!\pm\!6.8$ & $97.3\!\pm\!1.9$ & 55.6 \\
InCoM & $\bm{59.4\!\pm\!4.3}$ & $\bm{84.1\!\pm\!5.4}$ & $87.3\!\pm\!2.8$ & $\bm{84.1\!\pm\!4.6}$ & $82.5\!\pm\!5.6$ & $\bm{88.9\!\pm\!2.3}$ & $\bm{100\!\pm\!0.0}$ & $\bm{83.8}$ \\
\bottomrule
\end{tabular}
\end{small}
\end{center}
\end{table}

\begin{table}[ht]
\caption{TidyHouse and PrepareGroceries success rates (\%, mean$\pm$SD). Scenario means average the listed subtasks. $\ddagger$: pretrained robot policy, as in Table~\ref{table:set_table}.}
\label{table:tidy_prepare}
\small
\begin{center}
\begin{tabular}{lcccccc}
\toprule
& \multicolumn{3}{c}{TidyHouse} & \multicolumn{3}{c}{PrepareGroceries} \\
\cmidrule(lr){2-4} \cmidrule(lr){5-7} 
Method & Pick All & Place All & Mean & Pick All & Place All & Mean \\
\midrule
DP      & $0.0\pm0.0$ & $30.3\pm3.0$ & 15.2 & $0.4\pm0.3$ & $17.7\pm1.9$ & 9.1 \\
ACT     & $2.2\pm0.8$ & $31.6\pm3.2$ & 16.9 & $2.0\pm1.1$ & $27.5\pm2.9$ & 14.8 \\
WB-VIMA & $0.8\pm0.6$ & $29.4\pm3.5$ & 15.1 & $0.8\pm0.8$ & $22.0\pm2.6$ & 11.4 \\
DSPv2   & $1.3\pm0.3$ & $42.1\pm3.1$ & 21.7 & $0.9\pm0.6$ & $32.8\pm2.9$ & 16.9 \\
$\pi_0^{\ddagger}$ & $8.6\pm2.5$ & $31.1\pm3.0$ & 19.9 & $5.1\pm4.6$ & $27.2\pm3.7$ & 16.1 \\
InCoM & $\bm{16.7\pm1.9}$ & $\bm{78.9\pm2.9}$ & $\bm{47.8}$ & $\bm{15.0\pm2.4}$ & $\bm{65.9\pm3.1}$ & $\bm{40.5}$ \\
\bottomrule
\end{tabular}
\end{center}
\end{table}

\textbf{Evaluation Tasks.} We evaluate all methods on three representative mobile manipulation scenarios in ManiSkill-HAB: \textit{SetTable}, \textit{TidyHouse}, and \textit{PrepareGroceries}. These scenarios cover diverse task requirements including articulated object interaction, multi-task manipulation, and operation within constrained environments. Detailed task descriptions and visualization trajectories are provided in Appendix \ref{details_of_mshab}.

\textbf{Results Analysis.} Table~\ref{table:set_table} summarizes the results on the SetTable scenario, while Table~\ref{table:tidy_prepare} reports the results on the TidyHouse and PrepareGroceries scenarios. 

InCoM achieves the highest success rates on most tasks and consistently outperforms all baselines on average across the three scenarios, demonstrating strong mobile manipulation capability under perception-limited settings. Under matched training data and non-privileged observations, the gains over DSPv2 are 33.0/26.1/23.6 percentage points. DP, ACT, and DSPv2 exhibit notably lower success rates on pick-type tasks, suggesting limitations in precise object localization and fine-grained interaction when operating with dynamic viewpoints. WB-VIMA incorporates a hierarchical action decoder tailored for mobile manipulation; however, its performance remains constrained by insufficient perceptual modeling, particularly under dynamic attention requirements. Notably, AC-DiT, despite access to privileged state information, is still outperformed by InCoM overall, highlighting the effectiveness of intent-driven perception and coordinated action generation in challenging mobile manipulation scenarios.

Across the TidyHouse and PrepareGroceries scenarios, which involve multi-task manipulation, InCoM maintains substantially higher success rates than all baselines. In contrast, DP, ACT, and WB-VIMA show near-zero success rates on pick-all tasks, and DSPv2, while achieving moderate improvements in place-all tasks, still performs poorly on pick-all. These results indicate that InCoM maintains more stable performance as task complexity increases. The pretrained $\pi_0$ obtains 59.1/19.9/16.1\% across the three scenarios, compared with InCoM's 83.8/47.8/40.5\%.

\textbf{Trajectory Smoothness.} Aggregating the evaluated trajectories across all three ManiSkill-HAB scenarios, mean world-frame TCP spectral arc length~\cite{sparc} is $\mathbf{-4.65}$ for InCoM, $\mathbf{-6.18}$ for DSPv2, and $\mathbf{-6.27}$ for WB-VIMA. This dimensionless metric assigns higher (less negative) values to smoother motion, providing trajectory-level evidence complementary to task success rates.

\subsection{Ablation Study}
\label{exp:ablation}

\Needspace{13\baselineskip}
\begin{wraptable}{r}{0.40\textwidth}
\setlength{\abovecaptionskip}{0pt}
\setlength{\belowcaptionskip}{7pt}
\centering
\footnotesize
\caption{Component-level ablations in the \textit{SetTable} scenario.}
\label{tab:ablation_component}
\begin{tabular}{lcc}
\toprule
Method & Mean & $\Delta$ \\
\midrule
Full Model & $\bm{83.8}$ & -- \\
w/o IDPPM & 64.3 & $\downarrow$ 19.5 \\
w/o Scale Weights & 78.2 & $\downarrow$ 5.6 \\
w/o DARM & 77.6 & $\downarrow$ 6.2 \\
w/o DCFM & 75.9 & $\downarrow$ 7.9 \\
\bottomrule
\end{tabular}
\end{wraptable}

As summarized in Table~\ref{tab:ablation_component}, we first evaluate component-level ablations in the \textit{SetTable} scenario by replacing each component with a simpler or commonly used alternative while keeping the rest of the framework unchanged. Removing IDPPM and using a single-scale perceptual encoder leads to the largest performance drop, highlighting the critical role of multi-scale perception in handling stage-dependent sensing requirements for mobile manipulation. Replacing the adaptive scale weighting in IDPPM with a fixed multi-scale fusion results in a smaller but noticeable decrease, indicating that multi-scale features form the core perceptual representation, while intent-driven reweighting mainly optimizes their utilization across task stages. Substituting DARM with the Q-Former module from DSPv2~\cite{dspv2} consistently reduces performance, demonstrating that explicitly separating and refining geometric and semantic affinities yields more robust cross-modal alignment than unified query-based fusion. Finally, replacing DCFM with the dense head from DSPv2~\cite{dspv2} also degrades performance, confirming that modeling coordinated whole-body action generation via conditional flow matching is essential for stable base--arm coordination.

\Needspace{14\baselineskip}
\begin{wraptable}{r}{0.58\textwidth}
\setlength{\abovecaptionskip}{0pt}
\setlength{\belowcaptionskip}{7pt}
\centering
\footnotesize
\setlength{\tabcolsep}{4pt}
\caption{Fine-grained design ablations in the \textit{PrepareGroceries} scenario.}
\label{tab:ablation_finegrained}
\begin{tabular}{lcc}
\toprule
Variant & Mean & $\Delta$ \\
\midrule
Full Model & $\bm{40.5}$ & -- \\
History Encoder w/o Intent Modulation & 34.1 & $\downarrow$ 6.4 \\
Unidirectional Interaction & 36.1 & $\downarrow$ 4.4 \\
w/o Stop-Gradient & 37.2 & $\downarrow$ 3.3 \\
w/o Geometric Regularization & 38.1 & $\downarrow$ 2.4 \\
w/o Trend Token & 34.7 & $\downarrow$ 5.8 \\
\bottomrule
\end{tabular}
\end{wraptable}

We further conduct fine-grained design ablations in the \textit{PrepareGroceries} scenario. As shown in Table~\ref{tab:ablation_finegrained}, replacing intent modulation with a standard history encoder reduces the mean success rate from 40.5\% to 34.1\%, showing that historical actions are most effective when transformed into intent-driven perceptual scheduling. Unidirectional interaction also degrades performance, validating the importance of reciprocal base-arm coordination. Moreover, removing stop-gradient, geometric regularization, and the Trend Token consistently reduces success rates, confirming the effectiveness of gradient decoupling, geometry-guided alignment, and trajectory-level information exchange.

\subsection{Real-World Evaluation}
\label{exp:real_world}

\begin{figure}[htb] 
\centering
\resizebox{\textwidth}{!}{
\begin{tikzpicture}[x=1pt,y=1pt,font=\fontsize{7}{8}\selectfont]
\definecolor{actbar}{RGB}{172,185,207}
\definecolor{pibar}{RGB}{230,154,20}
\definecolor{incombar}{RGB}{0,139,102}
\foreach \panel/\heading in {0/Throw Rubbish,1/Close Drawer,2/Pick Banana,3/Move Block,4/Mean} {
  \begin{scope}[xshift=\panel*92pt]
    \foreach \tick in {0,20,40,60,80,100} {
      \draw[gray!22,line width=.25pt] (7,\tick*.72) -- (85,\tick*.72);
      \ifnum\panel=0
        \node[anchor=east,font=\fontsize{6}{7}\selectfont] at (5,\tick*.72) {\tick};
      \fi
    }
    \draw[gray,line width=.35pt] (7,0) rectangle (85,72);
    \node[font=\fontsize{7}{8}\selectfont\bfseries] at (46,89) {\heading};
  \end{scope}
}
\foreach \panel/\barx/\barvalue/\barlo/\barhi/\barcolor in {
0/21/5/0.8881/23.6131/actbar,0/46/15/5.2369/36.0419/pibar,0/71/50/29.9298/70.0702/incombar,
1/21/5/0.8881/23.6131/actbar,1/46/50/29.9298/70.0702/pibar,1/71/65/43.2854/81.8808/incombar,
2/21/10/2.7866/30.1034/actbar,2/46/35/18.1192/56.7146/pibar,2/71/45/25.8198/65.7915/incombar,
3/21/20/8.0658/41.6017/actbar,3/46/30/14.5477/51.8973/pibar,3/71/45/25.8198/65.7915/incombar,
4/21/10/5.1527/18.5107/actbar,4/46/32.5/23.2394/43.3595/pibar,4/71/51.25/40.4938/61.8916/incombar} {
  \begin{scope}[xshift=\panel*92pt]
    \fill[\barcolor] (\barx-7,0) rectangle (\barx+7,\barvalue*.72);
    \draw[black!65,line width=.45pt] (\barx,\barlo*.72) -- (\barx,\barhi*.72);
    \draw[black!65,line width=.45pt] (\barx-2,\barlo*.72) -- (\barx+2,\barlo*.72);
    \draw[black!65,line width=.45pt] (\barx-2,\barhi*.72) -- (\barx+2,\barhi*.72);
    \node[font=\fontsize{6}{7}\selectfont] at (\barx,\barhi*.72+5) {\barvalue};
  \end{scope}
}
\node[rotate=90,font=\fontsize{7}{8}\selectfont] at (-20,36) {Success rate (\%)};
\foreach \legendx/\barcolor/\legendname in {110/actbar/ACT,190/pibar/$\pi_{0.5}$,280/incombar/InCoM} {
  \fill[\barcolor] (\legendx,-15) rectangle (\legendx+8,-10);
  \node[anchor=west] at (\legendx+11,-12.5) {\legendname};
}
\end{tikzpicture}}
\caption{Real-world success rates with Wilson 95\% confidence intervals (20 trials per task; 80 trials for the pooled mean).}
\label{fig:real_world_success_rate}
\end{figure}

To further validate the effectiveness of InCoM in practical scenarios, we conduct real-world experiments on a Cobot-Magic robot platform across four representative mobile manipulation tasks: \emph{Throw Rubbish}, \emph{Close Drawer}, \emph{Pick Banana}, and \emph{Move Block}. For each task, we collect $150$ successful trajectories through manual teleoperation to construct the training dataset. We compare InCoM with two representative baselines: ACT~\cite{act} and the pretrained model $\pi_{0.5}$~\cite{pi05}. During evaluation, each method is executed for $20$ trials per task, and the final task success rate is reported. The detailed experimental setup and implementation details of the real-world experiments are provided in Appendix~\ref{details_of_real_world}.

\begin{figure}[!h] 
\centering
\includegraphics[width=\textwidth]{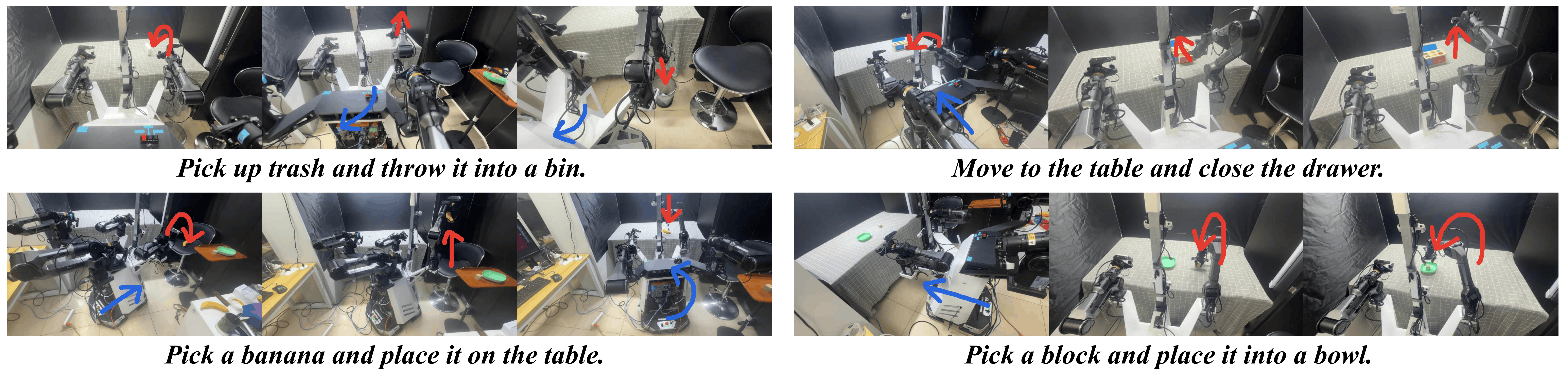}
\caption{Mobile manipulation demonstrations on the Cobot-Magic robot platform. Red arrows indicate the motion of the robotic arm, while blue arrows indicate the motion of the mobile base.}
\label{fig:real_world_vis_exp}
\end{figure}

Fig.~\ref{fig:real_world_success_rate} summarizes the quantitative comparison. 
InCoM consistently outperforms both baselines across all tasks, achieving a mean success rate of $51.25\%$, compared with $32.5\%$ for $\pi_{0.5}$ and $10\%$ for ACT. A task-stratified two-sided exact test against $\pi_{0.5}$ yields $\mathbf{p=0.0232}$, supporting a statistically significant aggregate advantage across these tasks; per-task confidence intervals and test details are provided in Appendix~\ref{details_of_real_world}. Qualitatively, baseline failures often involve insufficient base motion, whereas InCoM better coordinates base and arm to reach a feasible interaction pose. The results demonstrate that the proposed intent-driven perception and coordinated action generation enable more stable whole-body behaviors in real-world mobile manipulation scenarios.

\section{Conclusion and Limitations}
We present InCoM, an end-to-end framework for mobile manipulation that addresses two key challenges: stage-dependent perception under dynamic viewpoints and coordinated action generation between the mobile base and the manipulator. 
Extensive evaluations on SetTable, TidyHouse, and PrepareGroceries without privileged information, as well as real-world experiments, demonstrate that InCoM consistently outperforms existing methods, achieving higher task success rates and more stable whole-body motions in complex environments.

Despite strong performance, several limitations remain. First, training on task-specific data may limit generalization to unseen action patterns or object interactions; incorporating large-scale pretrained representations could improve robustness. Second, the current intent inference module may still struggle with complex or rapidly changing task dynamics. Finally, extending the framework to long-horizon tasks remains an important direction for future work.

\clearpage
\acknowledgments{This work was supported by the New Generation Artificial Intelligence-National Science and Technology Major Project 2025ZD0122902; in part by the National Natural Science Foundation of China (NSFC) under Grants 62136008 and 62293545; and by the Beijing Major Science and Technology Project under Contract no. Z251100008125023; and by Beijing Academy of Artificial Intelligence (BAAI).}

\bibliography{corl_2026}  

\newpage
\appendix
\onecolumn
\raggedbottom
\setlength{\intextsep}{8pt plus 2pt minus 2pt}
\setlength{\textfloatsep}{12pt plus 2pt minus 2pt}
\setlength{\floatsep}{10pt plus 2pt minus 2pt}

\section{Details of ManiSkill-HAB Experiments.}
\label{details_of_mshab}
\textbf{Robot Configuration.} We use the Fetch mobile manipulation robot as the agent. The robot is equipped with a 7-DoF manipulator, a vertically actuated torso, and a 2-DoF differential-drive mobile base.

\textbf{Observation Space.} To evaluate robustness under realistic sensing constraints, we modify the standard observation space of ManiSkill-HAB~\cite{mshab} by removing simulator-privileged signals. While the original environment provides high-precision state information such as target object pose and grasp indicators in addition to RGB, depth, and robot proprioception, such information is typically unavailable in real-world settings. Our setup excludes these privileged observations, requiring the agent to rely solely on visual inputs and basic proprioceptive feedback to infer object locations and grasp success.

\textbf{Action Space.} The agent's action space is a 13-dimensional continuous vector. Among these, 11 dimensions control joint position increments of the manipulator, torso, and head via PD joint controllers, while the remaining 2 dimensions control the linear and angular velocities of the mobile base. All action commands are normalized to the range $[-1, 1]$.

\textbf{Training Dataset.} We collect training data in ManiSkill-HAB using reinforcement learning algorithms. Specifically, PPO is used to collect trajectories for the open and close tasks, while SAC is used for the pick and place tasks. For each task, we collect 1,000 successful trajectories.

\textbf{Training Protocol.} For our reproduced non-VLA policies, we train each task on 1,000 trajectories for 100k training steps, using a batch size of 64, and report the task success rate as the evaluation metric. The pretrained $\pi_0$ is fine-tuned on the same task demonstrations, but has additional large-scale pretraining; pretrained visual backbones are also used by several non-VLA policies. Table~\ref{tab:hyperparameters} presents the hyperparameters used to train our method.

\textbf{Baseline Methods.} We compare InCoM with several representative mobile manipulation baselines. DP~\cite{dp} and ACT~\cite{act} adopt end-to-end policies with RGB and depth observations for joint action prediction, while DSPv2~\cite{dspv2} leverages RGB images and point cloud inputs. WB-VIMA~\cite{brs} employs a hierarchical action decoder designed for whole-body manipulation and uses relatively sparse colored point clouds as perceptual input. AC-DiT~\cite{ac-dit} uses RGB images, point clouds, and language instructions with a mobility-to-body conditioning mechanism. Its Table~1 results are reproduced here as reported numbers, not as our reimplementation: its code was unavailable when our paper was completed. Its privileged object/goal and EE-to-base poses and grasp indicator are explicitly distinguished in Table~\ref{table:set_table}. We additionally compare with the large-scale pretrained policy $\pi_0$.

\textbf{Evaluation Protocol.} For all policies evaluated by us, we use three evaluation seeds per checkpoint, with 189 episodes per task for each seed. Tables~\ref{table:set_table}--\ref{table:tidy_prepare} report the mean and standard deviation (SD) of each subtask's success rate across these three evaluation runs. AC-DiT's mean and SD are taken from its paper and follow its evaluation protocol. Scenario means average subtask means; subtask SDs alone do not determine the SD of a scenario mean.

{\setlength{\intextsep}{4pt}
\begin{table}[H]
\centering
\renewcommand{\arraystretch}{0.94}
\caption{Training hyperparameters.}
\label{tab:hyperparameters}
\begin{tabular}{lc}
\toprule
\textbf{Hyperparameter} & \textbf{Value} \\ 
\midrule
Optimizer Type                   & $Adam$ \\ 
Batch Size                       & 64 \\
Learning Rate                    & $1 \times 10^{-4}$ \\
Learning Rate Warm Up Steps      & 1000 \\
Learning Rate Cosine Decay Steps & 100,000 \\
$\lambda_{ent}$                  & 0.1 \\
$\lambda_{scale}$                & 1 \\
$\lambda_{align}$                & 0.01 \\
$\alpha$ (kinematic balancing)   & 0.375 \\
Sinkhorn iterations / $\epsilon$ & 5 / 0.05 \\
DCFM Euler steps at inference    & 1 \\

\bottomrule
\end{tabular}
\end{table}
}

\begin{figure}[H] 
\centering
\includegraphics[scale=0.28]{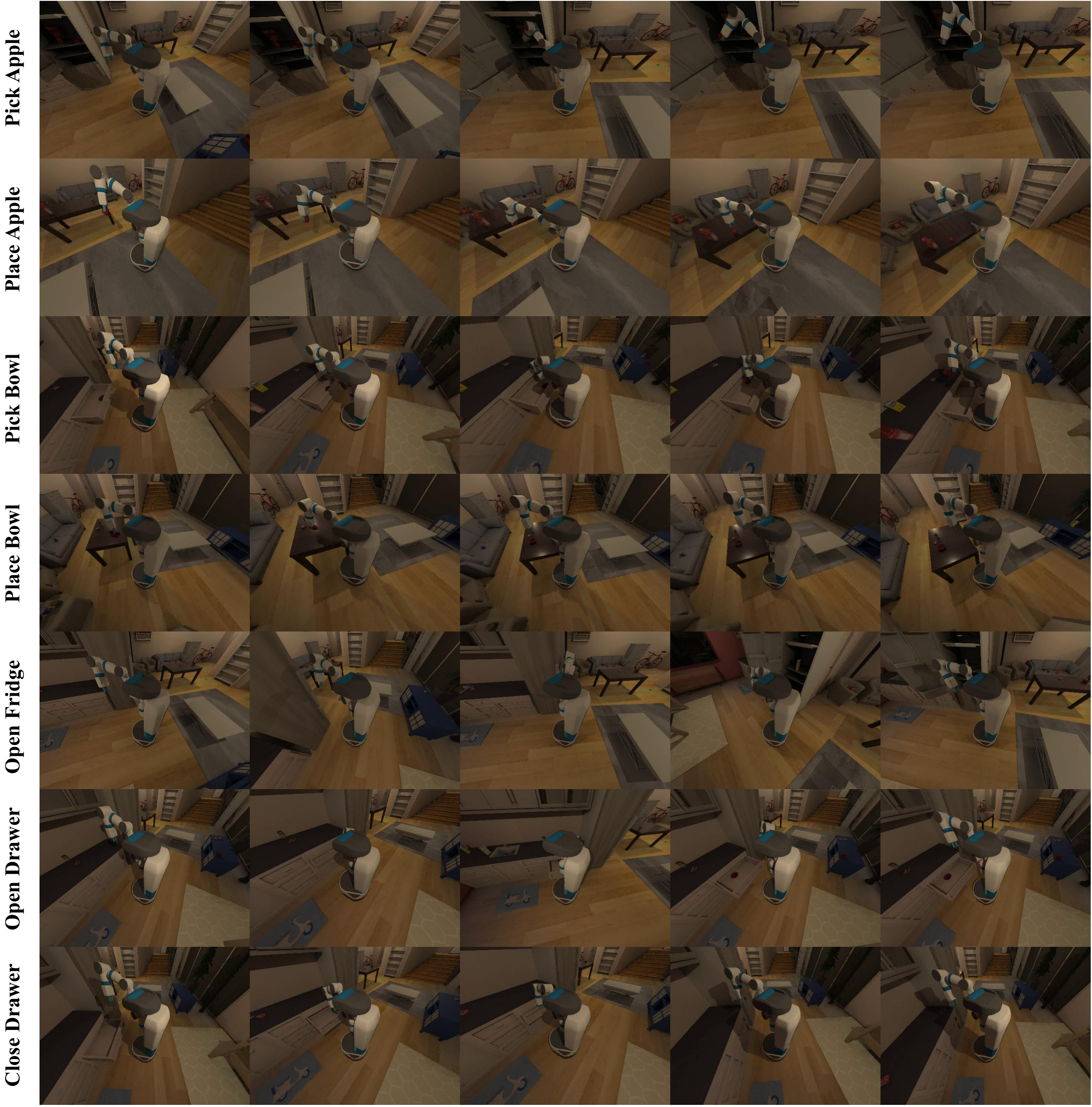}
\caption{Execution trajectories of our method across seven SetTable tasks.}
\label{fig:mshab_set_table_app}
\end{figure}

\begin{figure}[H] 
\centering
\includegraphics[scale=0.28]{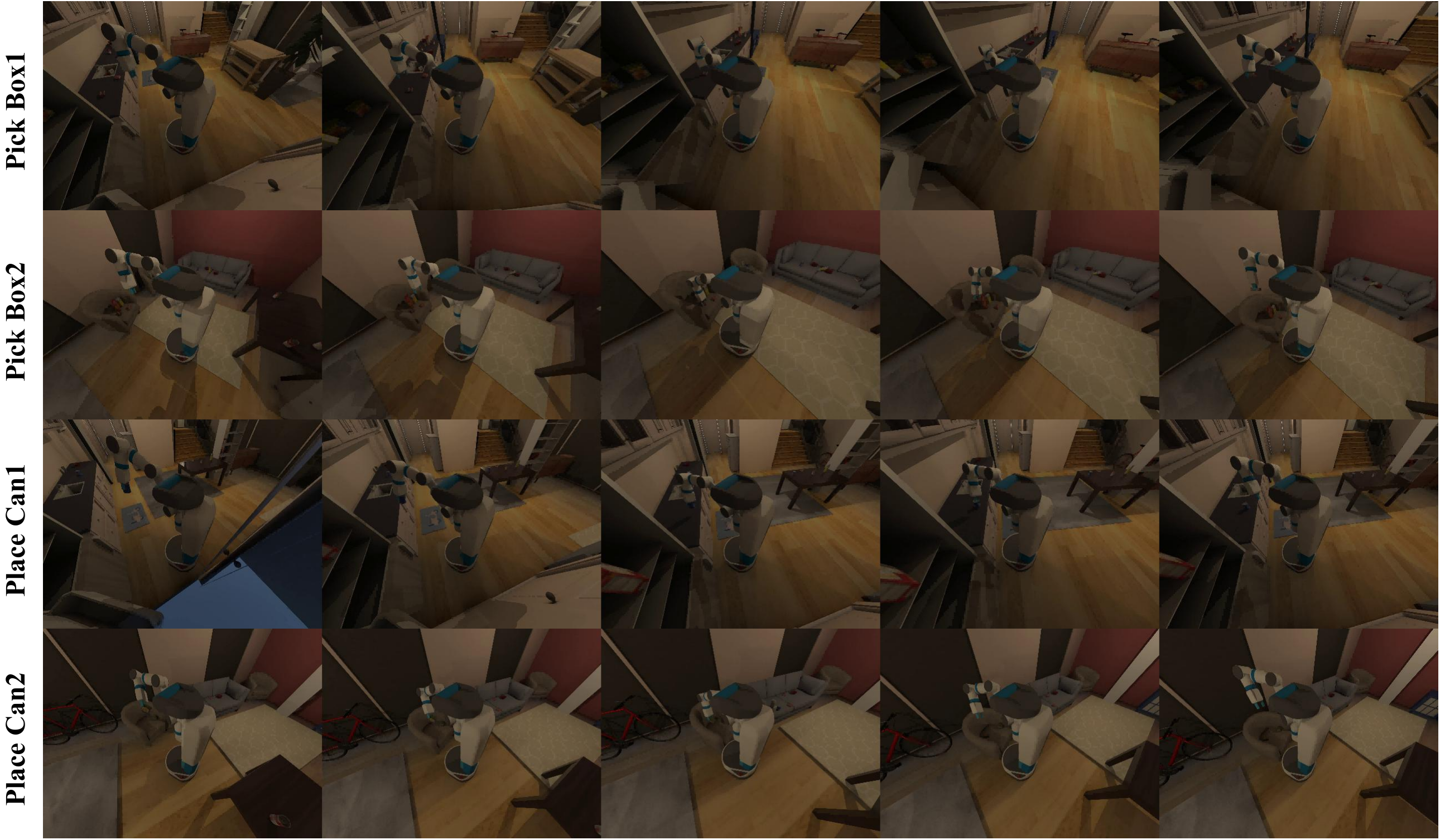}
\caption{Execution trajectories of our method across four TidyHouse and PrepareGroceries tasks.}
\label{fig:mshab_tidy_prep}
\end{figure}

\Needspace{0.97\textheight}
\textbf{Visualization of Task Execution Trajectories.} Figures \ref{fig:mshab_set_table_app} and \ref{fig:mshab_tidy_prep} visualize the execution trajectories of InCoM across multiple representative tasks.

\Needspace{0.94\textheight}
\begin{figure}[H] 
\centering
\includegraphics[scale=0.3]{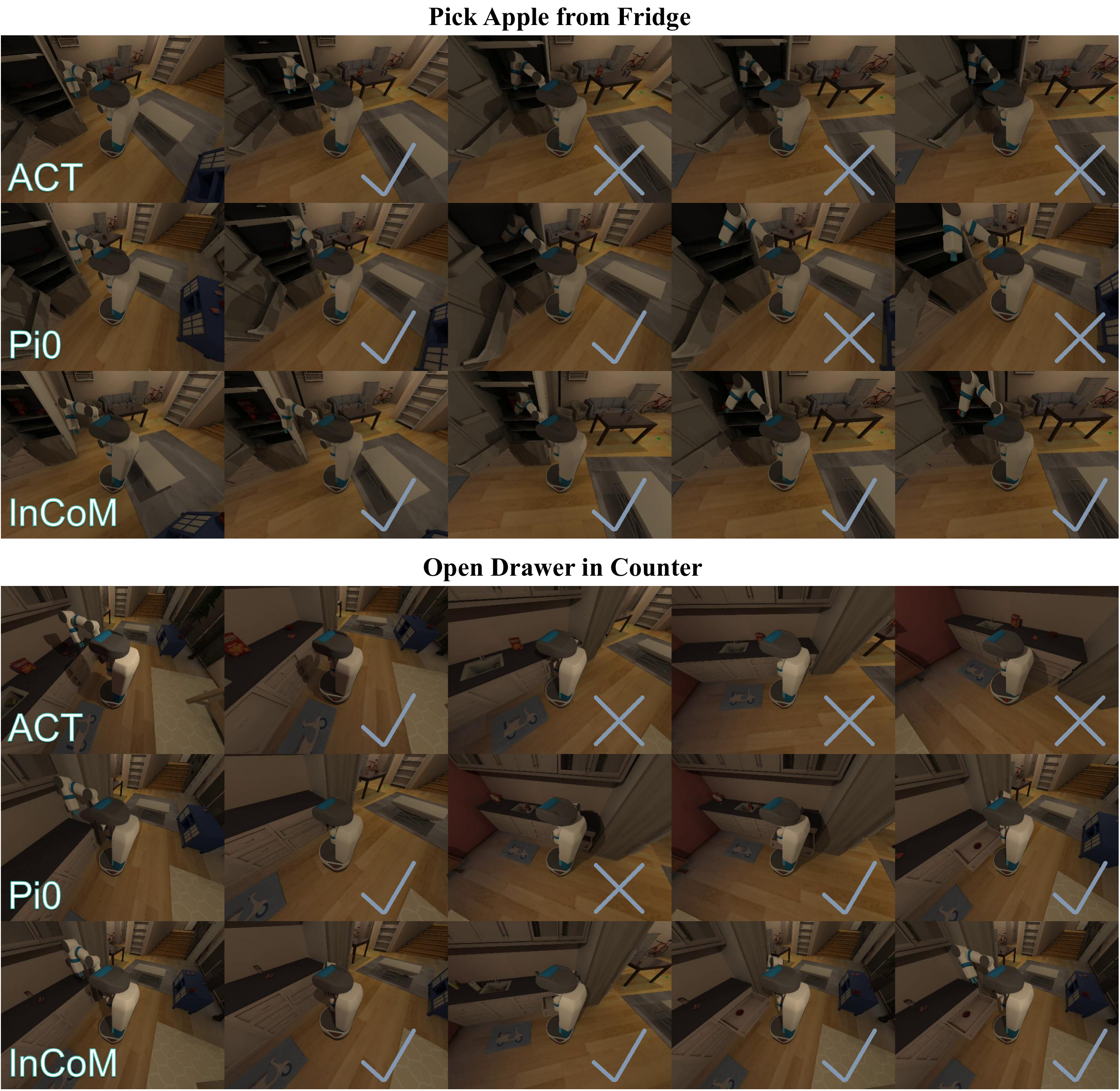}
\caption{Visualization of execution trajectories for ACT, $\pi_0$, and InCoM on the Pick Apple and Open Drawer tasks.}
\label{fig:contrast}
\end{figure}

\textbf{Comparative Analysis of Execution Trajectories.} We conduct a qualitative comparison of the execution trajectories of ACT, $\pi_0$, and InCoM on the Pick Apple and Open Drawer tasks, as shown in Fig.~\ref{fig:contrast}. In the Pick Apple task, ACT fails to effectively avoid the internal shelf inside the fridge due to limited perceptual capability, causing the manipulator to be repeatedly obstructed when approaching the target and preventing successful insertion into the fridge for grasping. $\pi_0$ manages to bypass the shelf, but fails to accurately localize the apple during the final interaction stage; as a result, the manipulator inadvertently pushes the apple toward the back of the fridge, leading to object drop and task failure. In contrast, InCoM is able to plan a reasonable manipulator trajectory under occlusion, precisely deliver the end-effector to the target region, and successfully grasp the apple. In the Open Drawer task, ACT exhibits insufficient coordination between the mobile base and the manipulator during the pulling motion. The base fails to appropriately support the arm movement, resulting in the drawer being opened only slightly. $\pi_0$ suffers from a similar issue in its first attempt, but eventually completes the task after a second adjustment. By comparison, InCoM opens the drawer in a single, continuous, and coordinated motion, demonstrating more stable and efficient base--arm coordination.

\begin{figure}[H] 
\centering
\includegraphics[scale=0.25]{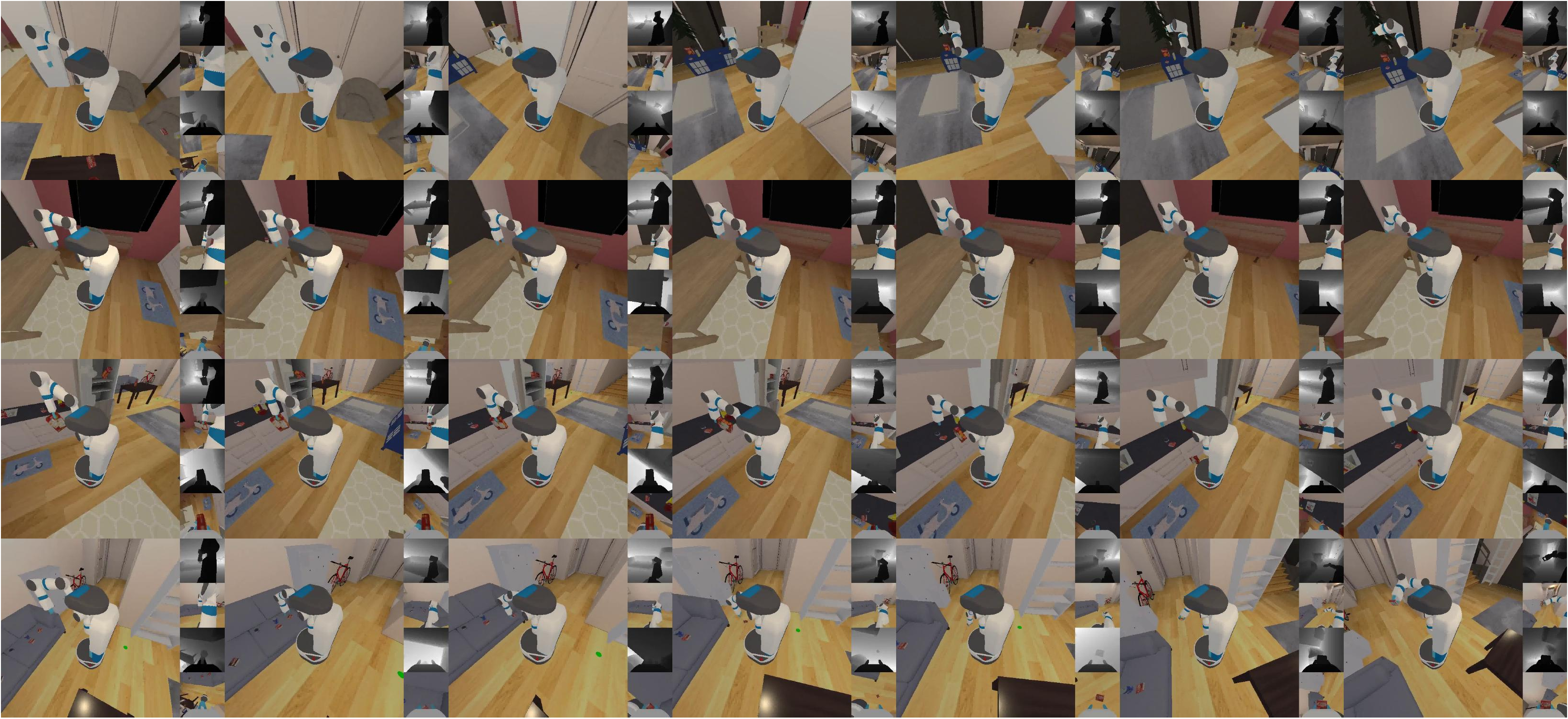}
\caption{\textbf{Representative failure cases of InCoM.} From top to bottom, the robot performs: picking an apple from a fridge, picking a can from a table, placing a box onto a countertop, and picking a box from a sofa.}
\label{fig:coner_case}
\end{figure}

\Needspace{0.69\textheight}
\textbf{Failure Case Analysis of InCoM.} We report several representative failure cases in Fig. \ref{fig:coner_case} to illustrate the current limitations of InCoM under challenging initial conditions and physical interactions. \textbf{(1) Target out of view due to initial placement.} In the task of picking an apple from the fridge, the robot was initialized at a lateral position relative to the fridge. As a result, the target object was not visible in the initial observations, preventing the policy from establishing a reliable perception--action plan and leading to task failure. \textbf{(2) Object instability at evaluation start.} In the task of picking a can from a table, the can was randomly spawned at an extreme edge of the tabletop. The object fell off immediately at the beginning of evaluation due to gravity, before any meaningful interaction could take place, resulting in an unavoidable failure.
\textbf{(3) Inaccurate size estimation during placement.} In the task of placing a box onto a countertop, the box instance was significantly larger than those commonly seen during training. The policy appeared to underestimate the object's spatial extent, causing the box to collide with the countertop edge during placement and subsequently fall. \textbf{(4) Unstable grasp with partial recovery.} In the task of grasping a box from a sofa, the initial grasp was unstable and the box slipped from the gripper onto the floor. However, the robot was able to subsequently re-detect the fallen object and successfully pick it up from the ground, demonstrating partial recovery capability despite the initial failure.

\Needspace{0.25\textheight}
\section{Multi-scale Modulation Weights Analysis}
\label{weight_analysis}
\begin{figure}[H] 
\centering
\includegraphics[width=\textwidth]{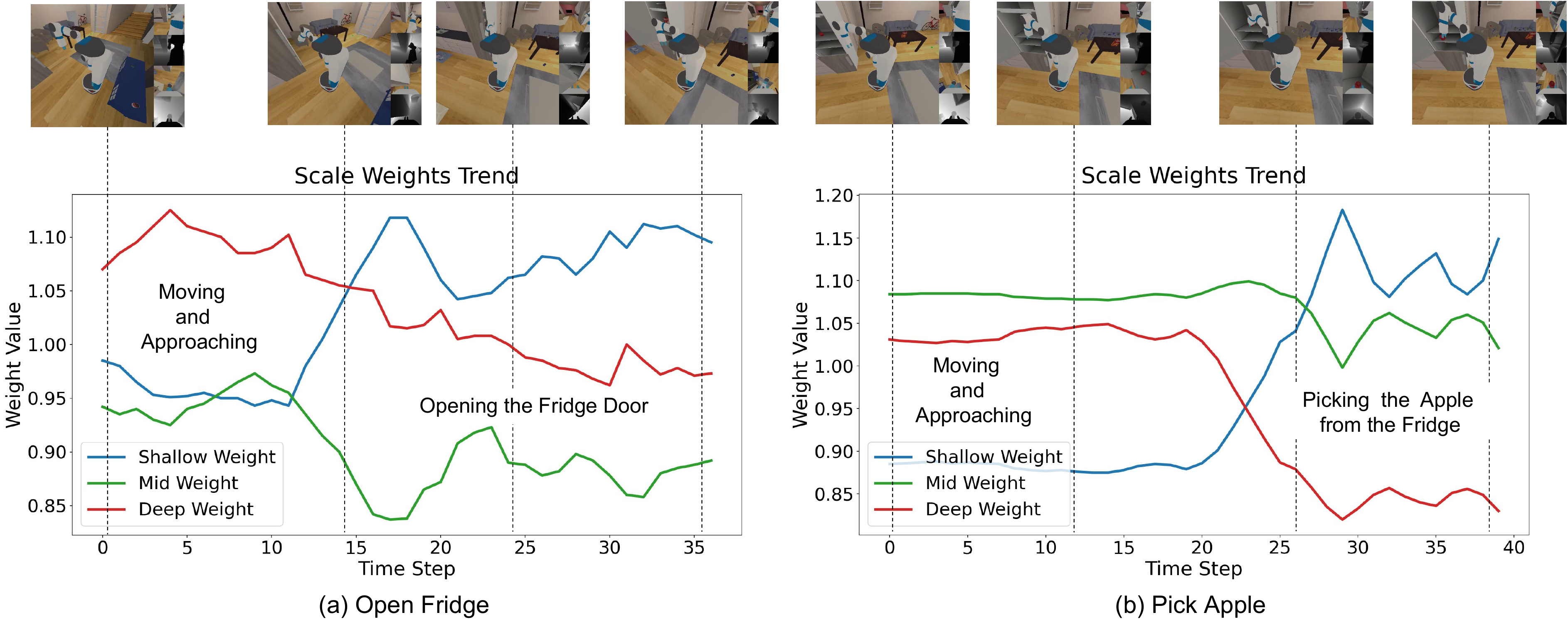}
\caption{IDPPM scale-weight evolution during execution. For visualization, the plotted weights are $3\,\mathrm{Softmax}(\phi(\mathbf{h}_t))$ (sum three); Eq.~(1) uses unscaled Softmax (sum one).}
\label{fig:mshab_scale_weight}
\end{figure}

Fig.~\ref{fig:mshab_scale_weight} illustrates the dynamic evolution of the three levels of modulation weights within the IDPPM during mobile manipulation task execution. By analyzing the temporal progression of these weights, we can intuitively observe how the system autonomously adjusts its perceptual focus in response to latent motion intent. For visibility, the figure plots $3\,\mathrm{Softmax}(\phi(\mathbf{h}_t))$, whose components sum to three; the model uses the unscaled normalized weights in Eq.~(1).

\textbf{Sensitivity to Kinematic Balancing.} We choose $\alpha=0.375$ to put arm/base activity on comparable scales despite their different dimensionalities and magnitudes. This setting yields the expected stage-dependent allocation across shallow, middle, and deep features. With $\alpha=1$, the qualitative allocation pattern remains, but the training success curve fluctuates more and reaches a lower peak. We choose $\lambda_{scale}=1$ and $\lambda_{align}=0.01$ by balancing loss magnitudes and keep them fixed across tasks.

\Needspace{0.55\textheight}
As shown in Fig. \ref{fig:mshab_scale_weight}(a,b), during the early stage of the task, the robot mainly executes base motions to approach the target, where deep-level features receive higher weights. This aligns with the design intuition that global scene representations are more critical at this stage for safe navigation and coarse motion planning. As the robot nears the target and transitions into the grasping phase, the weights progressively shift toward shallow-level features, indicating a change in perceptual focus. This behavior shows that IDPPM increasingly emphasizes fine-grained local details, which are essential for precise end-effector control, such as object boundary alignment and contact reasoning.

\Needspace{0.48\textheight}
\section{Details of Real-World Experiments.}
\label{details_of_real_world}

\begin{figure}[H] 
\centering
\includegraphics[width=\textwidth]{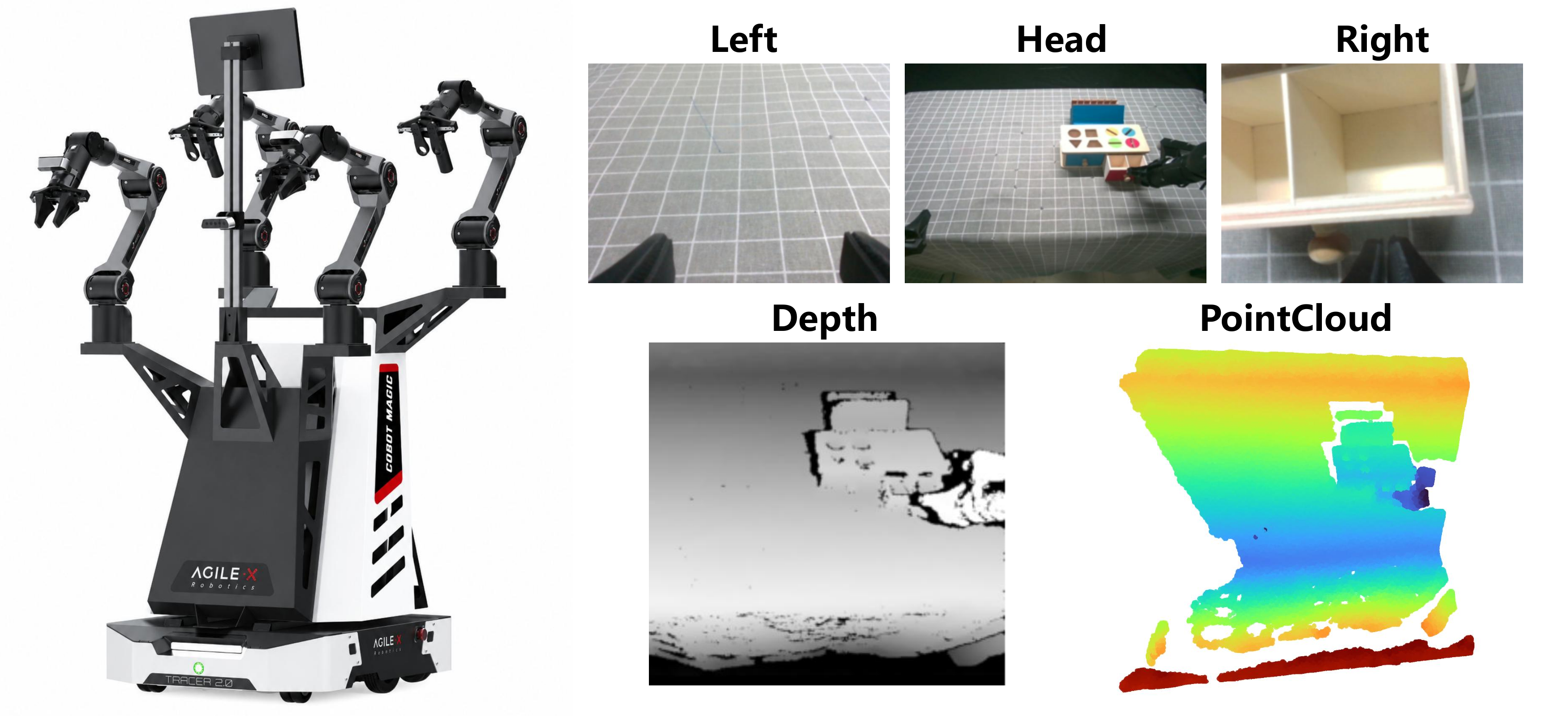}
\caption{Overview of the mobile robot platform and its multimodal observations used in our experiments, including multi-view RGB images, depth maps, and 3D point clouds.}
\label{fig:obs_vis}
\end{figure}

\textbf{Robot Platform.}
Our real-world experiments are conducted on the Cobot-Magic (Fig.~\ref{fig:obs_vis}) mobile manipulation platform developed by AgileX Robotics. The robot consists of a mobile base and two 7-DoF manipulators. Each manipulator is controlled in joint space with a 7-dimensional action vector corresponding to the joint commands. The mobile base is controlled via velocity commands, including linear and angular velocities.

\begin{figure}[H] 
\centering
\includegraphics[width=\textwidth]{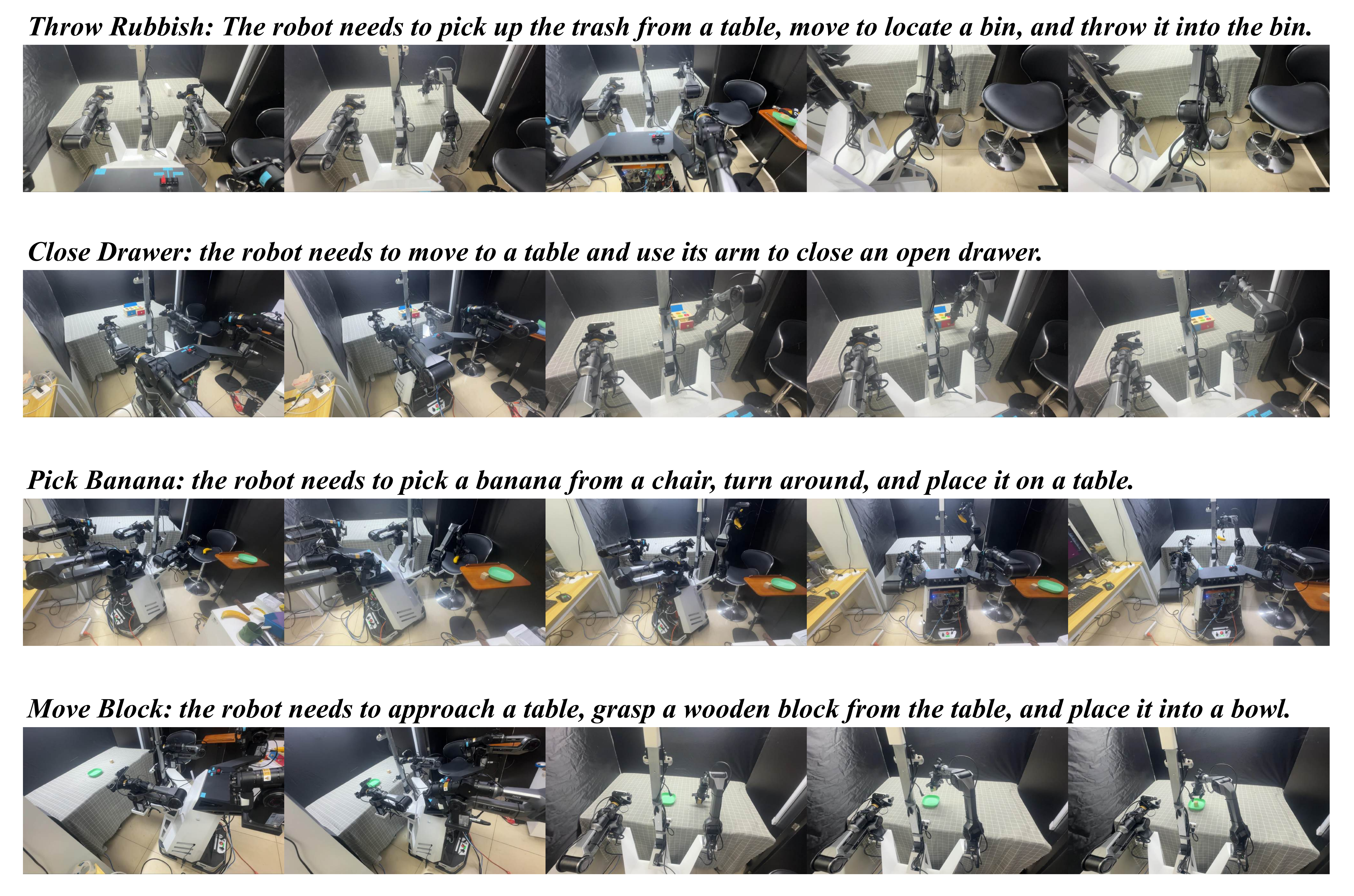}
\caption{Execution snapshots of the robot performing real-world mobile manipulation tasks using our InCoM policy, where each row corresponds to a different task. The images illustrate key stages of task execution from left to right.}
\label{fig:real_world_vis_traj}
\end{figure}

\Needspace{0.55\textheight}
\textbf{Execution Snapshots.}
To provide a clearer understanding of the robot behavior in real-world experiments, we present execution snapshots of our InCoM policy during task execution, as shown in Fig.~\ref{fig:real_world_vis_traj}. The images capture several key stages of the mobile manipulation process, including moving toward the target workspace, object interaction, and task completion.

\textbf{Training Setup.}
Demonstrations are collected by two synchronized operators: one controls the base while the other controls both arms through a leader--follower interface. This permits simultaneous base--arm motion, rather than imposing a base-then-arm sequence. Each method uses the same 150 task-specific demonstrations. Simulation and real-world policies are trained separately; these experiments do not evaluate sim-to-real transfer.

For the ACT baseline and InCoM, we use a batch size of $64$ and train the models for $50$ epochs. 
For the $\pi_{0.5}$ baseline, training is conducted for $30{,}000$ steps with a batch size of $128$. 
Following the standard fine-tuning practice for VLA, we apply LoRA to the VLM backbone while fully fine-tuning the action prediction head. 
All training is performed on NVIDIA A100 servers.

\textbf{Inference Setup.}
During inference, ACT predicts a future action horizon of $32$ steps and applies temporal ensembling of overlapping predictions to generate the final control commands. The $\pi_{0.5}$ baseline adopts an action chunk size of $25$, which performed better than $50$ and $8$ in our validation. InCoM maintains the same inference configuration as used in the ManiSkill-HAB simulation experiments, with a prediction horizon of $T_p=8$. These are complete-policy comparisons using official or best-performing configurations, not horizon-controlled architectural ablations.

\textbf{Evaluation Setup.}
For each task, we conduct $20$ independent trials to evaluate task success rates. 
In every trial, both the initial pose of the robot and the position of the target object are randomly perturbed within a small range, using matched initialization ranges, object placements, lighting, and backgrounds across methods. Success requires completing the full navigation--manipulation workflow.

\begin{table}[H]
\centering
\small
\caption{Real-world success rates (\%) with Wilson 95\% confidence intervals, based on 20 trials per task. The mean equally weights the four tasks.}
\label{tab:real_world_ci}
\begin{tabular}{lccccc}
\toprule
Method & \shortstack{Throw\\Rubbish} & \shortstack{Close\\Drawer} & \shortstack{Pick\\Banana} & \shortstack{Move\\Block} & Mean \\
\midrule
ACT & $5\,[1,24]$ & $5\,[1,24]$ & $10\,[3,30]$ & $20\,[8,42]$ & $10.00\,[5,19]$ \\
$\pi_{0.5}$ & $15\,[5,36]$ & $50\,[30,70]$ & $35\,[18,57]$ & $30\,[15,52]$ & $32.50\,[23,43]$ \\
InCoM & $\bm{50}\,[30,70]$ & $\bm{65}\,[43,82]$ & $\bm{45}\,[26,66]$ & $\bm{45}\,[26,66]$ & $\bm{51.25}\,[40,62]$ \\
\bottomrule
\end{tabular}
\end{table}

\textbf{Uncertainty and Statistical Comparison.} Table~\ref{tab:real_world_ci} gives per-task success rates and Wilson intervals. Across the four fixed tasks, InCoM succeeds in 41/80 trials and $\pi_{0.5}$ in 26/80. We condition on the total number of successes within each task and convolve the four hypergeometric null distributions. The two-sided probability of an aggregate difference at least as large as observed is $p=0.0232$. This unpaired, task-stratified exact test supports an aggregate advantage on these evaluated tasks; it does not establish per-task significance or uncertainty from training randomness. Qualitatively, ACT and $\pi_{0.5}$ failures often involve insufficient base motion, whereas InCoM better coordinates base and arm to reach a feasible interaction pose.

\Needspace{0.35\textheight}
\section{Policy Implementation Details and Efficiency Analysis}
\label{policy_details}
\textbf{Policy Implementation Details.} We provide additional policy implementation details that were omitted in the main manuscript. 

\textbf{DARM Tensor Operations and Cost.} At shallow/middle/deep scales, we bound the voxel-token counts by $N=256/128/64$. Each camera contributes a $16\times16$ image grid, so $M=256V$ for $V$ views ($M=512$ in the two-view simulation setting). With eight attention heads, both affinity maps have shape $B\times8\times N\times M$. Batched matrix products avoid an $N\times M\times D$ pairwise feature tensor. The grouped $1\times1$ refiner maps 16 input channels to 64 hidden channels and then eight output channels, with eight groups and 264 parameters. The resulting patch-wise attention aggregates image values into voxel tokens.

The pairwise affinity and aggregation cost is $O(\sum_i N_iMD)$, where $D=512$ is the total feature dimension and $i$ indexes scales. Its dependence on $NM$ remains dense; token bounding prevents growth with the raw point count, but the cost still grows with the number of views. For two views in fp32, the two base maps occupy 8/4/2\,MiB per sample at the three scales; the largest refiner intermediate is 32\,MiB per sample. These figures describe these tensors only, not total training memory or all temporary allocations.

The full model has approximately 150M parameters, of which 60M are trainable. All reported inference results use one Euler step.

We use the DINOv2-base model as the pretrained vision backbone. For fine-tuning, we apply rsLoRA~\cite{rslora} to all linear layers, with both the rank $r$ and scaling factor $\alpha_{\mathrm{LoRA}}$ set to 16, and a dropout rate of 0.1. This scaling factor is distinct from the kinematic balancing coefficient in Eq.~(2). For multi-scale feature extraction, we use intermediate representations from the 4th, 8th, and 12th layers of DINOv2.

The History Transformer adopts an architecture with 2 layers and 4 attention heads, 
an embedding dimension of 256, and a historical window size of $H=8$.

\begin{figure}[H]
\centering
\includegraphics[scale=0.45]{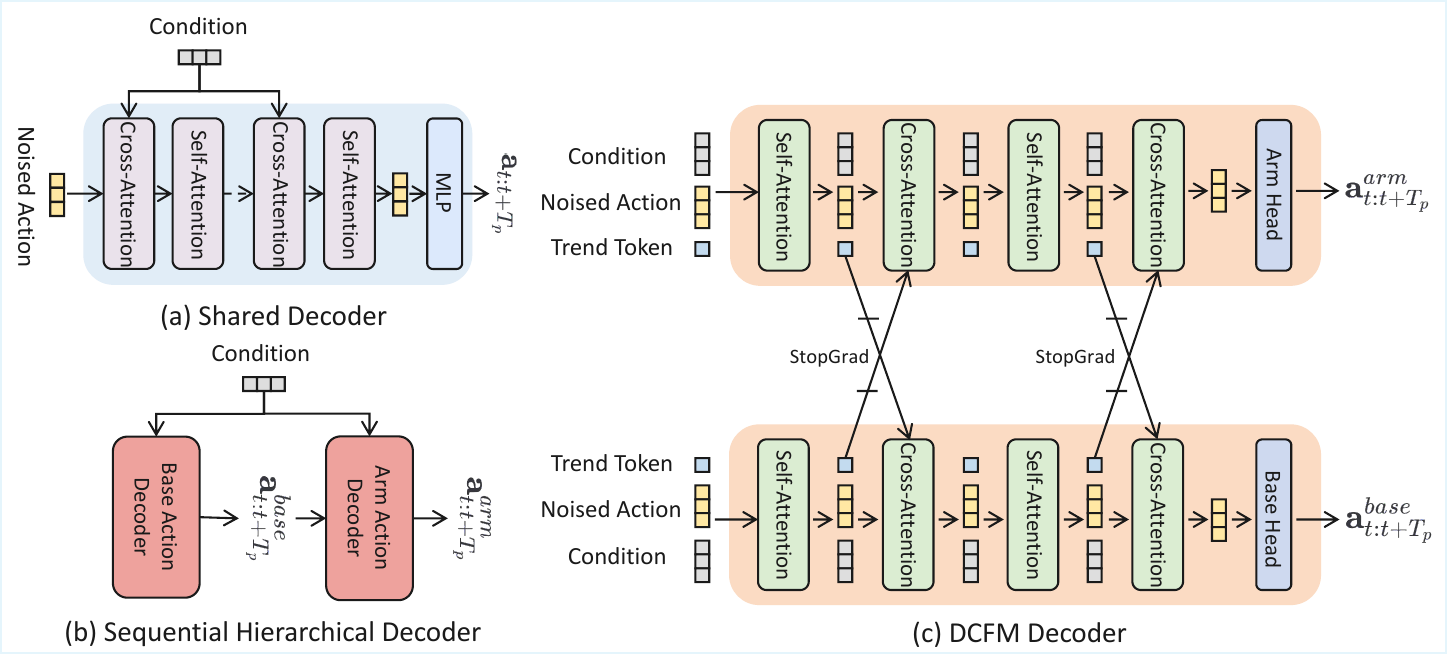}
\caption{\textbf{Comparison of the Action Decoders.} (a) Shared Decoder: base and arm actions are jointly modeled by a single decoder with a unified output head. (b) Sequential Hierarchical Decoder: base and arm actions are predicted by independent decoders, where the arm decoder is conditioned on the base output, capturing only unidirectional dependency. (c) DCFM Decoder: base and arm actions are decoded in parallel and exchange information bidirectionally via cross-attention in intermediate layers. To avoid unstable gradient interference during training, stop-gradient is applied in the cross-attention, retaining only forward conditional information.}
\label{fig:action_head}
\end{figure}

\Needspace{0.53\textheight}
In DCFM, to implement bidirectional coordination between the mobile base and the manipulator,
we employ two parallel Transformer decoder branches. Each branch maintains a learnable Trend Token that aggregates motion trends
from the predicted action sequence. The hidden dimension of the decoder is set to $256$, with $8$ attention heads per layer, and the decoder consists of $4$ stacked Transformer-based decoder blocks. The action prediction horizon is set to $T_p=8$.
Within each block, both branches first perform self-attention to model intra-trajectory dependencies. After that, cross-attention is applied between the two branches with stop-gradient to stabilize training while preserving bidirectional coordination. The overall architecture of the action decoder is illustrated in Fig.~\ref{fig:action_head}.

\Needspace{0.33\textheight}
\textbf{Inference Efficiency.}
We additionally evaluate the inference time and GPU memory consumption of different methods. All tests are conducted on an NVIDIA A100 GPU under the same environment settings. Table~\ref{tab:efficiency_diffusion} reports the results for diffusion-based methods, while Table~\ref{tab:efficiency_nondiffusion} presents the comparison for non-diffusion-based methods.

\begin{table}[H]
\centering
\caption{Inference efficiency comparison for diffusion-based methods.}
\label{tab:efficiency_diffusion}
\begin{tabular}{lccc}
\toprule
Metric & DP~\cite{dp} & WB-VIMA~\cite{brs} & InCoM (Ours) \\
\midrule
Inference Time & 790 ms & 370 ms & 140 ms \\
GPU Memory & 3.7 GB & 3.8 GB & 5.4 GB \\
\bottomrule
\end{tabular}
\end{table}

\begin{table}[H]
\centering
\caption{Inference efficiency comparison for non-diffusion-based methods.}
\label{tab:efficiency_nondiffusion}
\begin{tabular}{lcc}
\toprule
Metric & ACT~\cite{act} & DSPv2~\cite{dspv2} \\
\midrule
Inference Time & 20 ms & 70 ms \\
GPU Memory & 3.8 GB & 5.4 GB \\
\bottomrule
\end{tabular}
\end{table}

As shown in the tables, InCoM achieves significantly faster inference than the compared diffusion-based baselines, reducing latency from 790 ms in DP and 370 ms in WB-VIMA to 140 ms. Although the GPU memory usage of InCoM is moderately higher due to the multi-scale perception modeling, it remains within a practical range for modern GPUs.

Furthermore, based on our real-world experiments, the inference latency of InCoM on an 
RTX 4090 GPU is approximately 60 ms, enabling real-time deployment and inference.

\Needspace{0.97\textheight}
\section{Visualization Analysis of Feature Receptive Field and Multi-scale Spatial Attention}

\begin{figure}[H] 
\centering
\includegraphics[scale=0.29]{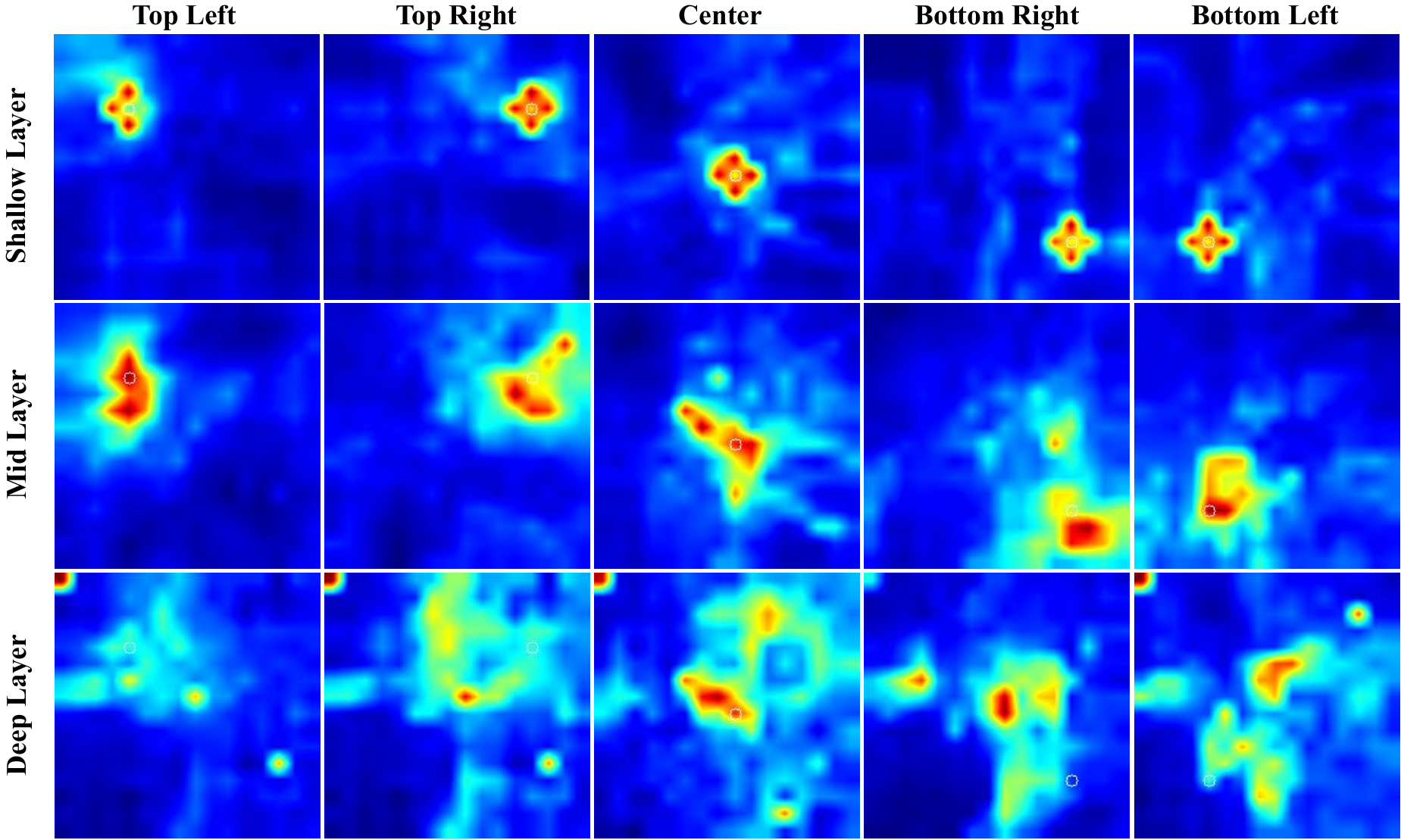}
\caption{\textbf{Visualization of patch-to-patch attention across network depths in InCoM.} Each row corresponds to a different encoder layer (shallow, mid, deep), and each column shows one of five representative query patches. White hollow circles indicate query patch centers, and the heatmap color represents attention strength. Shallow features focus on local details, mid-layer features capture broader context, and deep features integrate global scene information, forming a multi-scale perceptual hierarchy.}
\label{fig:attention_scale}
\end{figure}

To further investigate how the image encoder in InCoM processes features at different network depths, we conduct a visualization analysis of its receptive fields. Specifically, we apply a patch-to-patch attention analysis method, selecting five representative spatial locations in the input image as query patches to examine how attention is distributed across the multi-scale feature hierarchy. The selected query patches include the geometric center of the image and four positions near the corners while avoiding the image boundaries, minimizing potential edge effects inherent to the Vision Transformer architecture. The visualization is presented as a 3\ensuremath{\times}5 heatmap matrix, with rows corresponding to different encoder layers (shallow, mid, and deep from top to bottom) and columns representing the five query patches. White hollow circles indicate the centers of the query patches, and the heatmap color reflects attention strength, with red regions corresponding to areas contributing most to the query patch.

As shown in Fig.~\ref{fig:attention_scale}, the attention maps reveal a clear hierarchy in spatial modeling across different depths. In shallow layers, attention is concentrated around the query patch, indicating strong locality and a focus on capturing fine-grained details. In mid layers, attention gradually expands to cover larger contextual regions related to the query, demonstrating the model's capacity to capture broader scene context. In deep layers, attention extends even further, forming a pronounced global receptive field and enabling the integration of information across the entire scene. It is worth noting that a persistent high-response region in the upper-left corner of the deep-layer heatmaps corresponds to the commonly observed ``Attention Sink'' in Vision Transformers~\cite{visiontransformersneedregisters, attentionsinkemergeslanguage, sinksinkvisualinformation}, which serves as a global register for storing scene-level information and does not interfere with perception of actual physical targets. Because the Attention Sink absorbs a fixed proportion of absolute attention weights, the attention assigned to actual physical targets and surrounding background appears relatively smaller; during linear normalization of the heatmaps, these low-amplitude but semantically rich responses are visually suppressed, making the deep-layer receptive field appear narrower than it truly is. In reality, deep features implicitly encompass a broader and more coherent spatial representation.

Overall, this visualization analysis demonstrates that the image encoder used in InCoM produces complementary representations at different depths: shallow features retain high-resolution local details, while deep features provide high-level semantic information with global scene context. This multi-scale feature modeling establishes a stable and rich perceptual foundation for subsequent intent-driven modulation and coordinated whole-body action generation.

\end{document}